\def\year{2022}\relax
\documentclass[letterpaper]{article} %
\usepackage{aaai22}  %
\usepackage{times}  %
\usepackage{helvet}  %
\usepackage{courier}  %
\usepackage[hyphens]{url}  %
\usepackage{graphicx} %
\usepackage{natbib}  %
\usepackage{caption} %
\DeclareCaptionStyle{ruled}{labelfont=normalfont,labelsep=colon,strut=off} %
\usepackage[switch]{lineno}
\usepackage{bm}
\usepackage{amssymb}
\usepackage{subfigure}
\usepackage{array}
\usepackage{algorithmic}

\usepackage{amsmath}
\usepackage{etoolbox}
\newcommand*\linenomathpatch[1]{%
  \cspreto{#1}{\linenomath}%
  \cspreto{#1*}{\linenomath}%
  \csappto{end#1}{\endlinenomath}%
  \csappto{end#1*}{\endlinenomath}%
}

\linenomathpatch{equation}
\linenomathpatch{gather}
\linenomathpatch{multline}
\linenomathpatch{align}
\linenomathpatch{alignat}
\linenomathpatch{flalign}

\usepackage[linesnumbered,ruled,vlined]{algorithm2e}
\usepackage{multirow}

\title{Learning to Learn Transferable Attack}
\author{
  Shuman Fang$^{\dag*}$,
  Jie Li$^{\dag*}$,
  Xianming Lin$^{\dag}$,
  Rongrong Ji$^{\dag\ddagger\natural}$
}
\affiliations{
  $^{\dag}$Media Analytics and Computing Lab, Department of Artificial Intelligence, \\School of Informatics, Xiamen University, China \\
  $^{\ddagger}$Peng Cheng Laboratory, Shenzhen, China \\
  fangshuman@stu.xmu.edu.cn, lijie.32@outlook.com, linxm@xmu.edu.cn, rrji@xmu.edu.cn \\
  $^*$Contributed Equally, $^\natural$Corresponding Author
}

\begin{document}
\maketitle

\begin{abstract}

Transfer adversarial attack is a non-trivial black-box adversarial attack that aims to craft adversarial perturbations on the surrogate model and then apply such perturbations to the victim model. However, the transferability of perturbations from existing methods is still limited, since the adversarial perturbations are easily overfitting with a single surrogate model and specific data pattern. In this paper, we propose a \emph{Learning to Learn Transferable Attack (LLTA)} method, which makes the adversarial perturbations more generalized via learning from both data and model augmentation. For data augmentation, we adopt simple random resizing and padding. For model augmentation, we randomly alter the back propagation instead of the forward propagation to eliminate the effect on the model prediction. By treating the attack of both specific data and a modified model as a task, we expect the adversarial perturbations to adopt enough tasks for generalization. To this end, the meta-learning algorithm is further introduced during the iteration of perturbation generation. Empirical results on the widely-used dataset demonstrate the effectiveness of our attack method with a 12.85\% higher success rate of transfer attack compared with the state-of-the-art methods. We also evaluate our method on the real-world online system, \emph{i.e.}, Google Cloud Vision API, to further show the practical potentials of our method.

\end{abstract}

\section{Introduction}

Though showing dominant advance in various tasks, deep neural networks (DNNs) are proven vulnerable to adversarial examples~\cite{goodfellow2015explaining},
\emph{i.e.}, adding well-designed human-imperceptible perturbations into natural images can mislead DNNs.
Such a drawback has raised high security concerns when deploying DNNs in security-sensitive scenarios~\cite{Xiao_2021_CVPR,fang2021invisible}, which has caused has caused researchers to attach attention to model security~\cite{wu2020adversarial,naseer2020self}.
According to the knowledge of the victim model, the adversarial attack can be categorized into either white-box or black-box attack.
The black-box attack is more practical but challenging, and can be further divided into the query-based attack and transfer-based attack.
The query-based attack polishes the adversarial perturbations by querying the victim model iteratively, which achieves a high attack success rate but brings more resource consumption~\cite{lilyas2018prior, li2020projection,li2021aha}.
Instead,
the transfer-based attack crafts the adversarial perturbations on a white-box surrogate model and then transfer to the victim model directly, which is more efficient and avoids suspicion.
Essentially, the adversarial transferability is highly expected since the surrogate model should be ideally similar to the victim model~\cite{orekondy_knockoff_2019,wang2021black}.
However, as in reality the perturbations usually overfit to the surrogate model, the transfer success rate remains unsatisfactory.

\begin{figure}[t]
    \centering
    \includegraphics[width=0.48\textwidth]{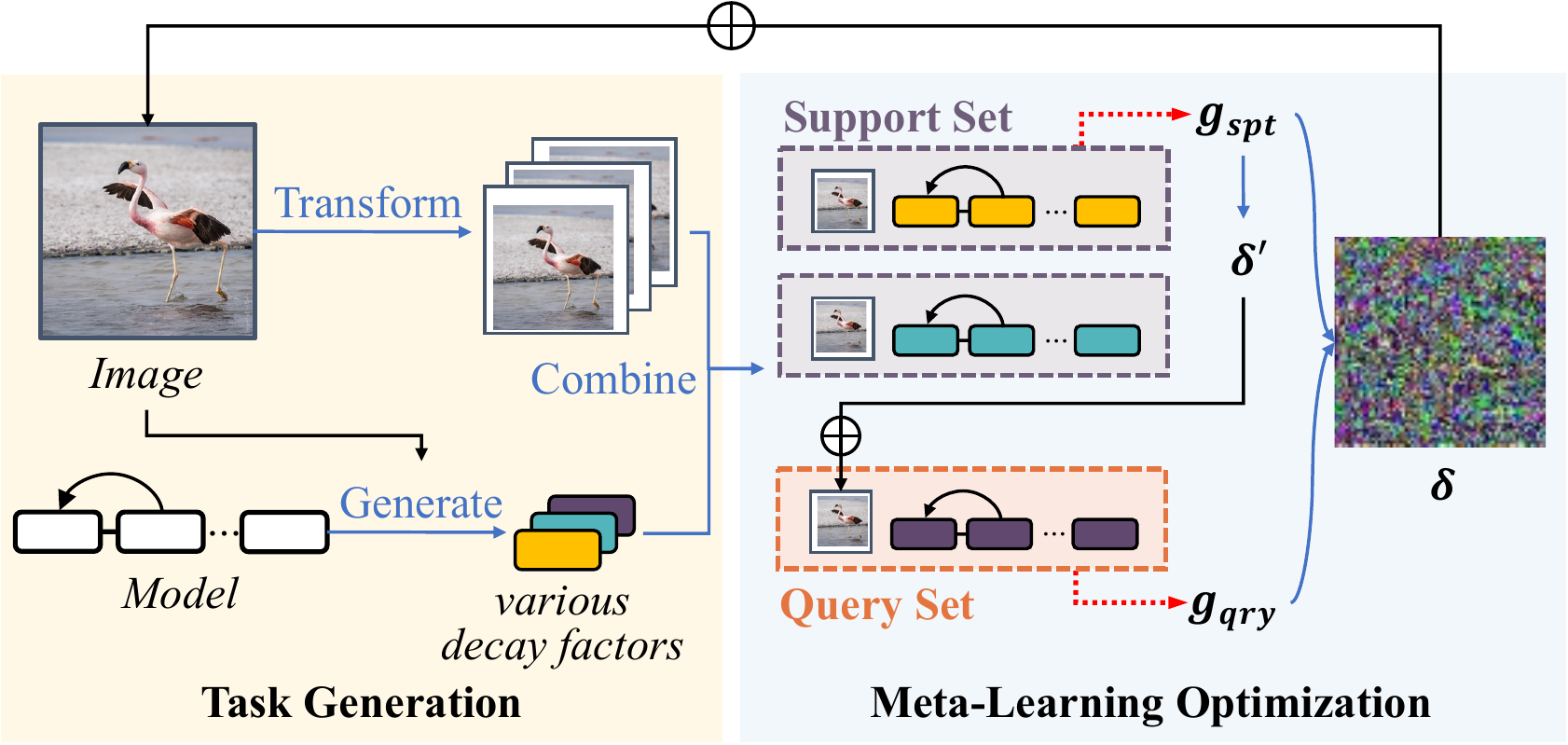}
    \caption{Overview of Learning to Learn Transferable Attack, which includes task generation and meta-learning optimization. \textit{In task generation}, treating attacking of an augmented data and model as a task, we generate multiple tasks by adopting random transformation (data augmentation) and introducing a group of learnable decay factors on back propagation process (model augmentation). \textit{In meta-learning optimization}, we split these tasks into support set and query set, craft a temporary perturbation from the support set and finetune it on the query set. Finally, we update the actual perturbation with the gradient from two sets.}
    \label{fig:overview}
\end{figure}

To this end, many efforts have been put to learn more generalized perturbations.
\citet{dong2018boosting} and \citet{lin2019nesterov} integrated optimization techniques.
These techniques are effective but with limited improvement space for adversarial attacks due to the small number of optimization steps here.
\citet{huang2019enhancing} and \citet{2538_ECCV_2020_paper} only perturbed features from the intermediate layer of DNNs,
but the layer should be picked manually and carefully.
There are also some augmentation-based methods~\cite{xie2019improving,lin2019nesterov,liu2017delving,2004.05790v1}.
These methods narrow the gap between the surrogate model and the victim model to some degree,
but still suffer from insufficient augmentation or extra costs, \emph{e.g.}, multiple surrogate models for the model ensemble.

In this paper, we propose a novel transfer adversarial attack framework, termed \emph{Learning to Learn Transferable Attack} (LLTA), to enhance the transferability of adversarial perturbations.
As depicted in Fig.\,\ref{fig:overview}, our method contains a task generation module based on both data and model augmentation, and the meta-learning optimization module to finally update the perturbations.
For data augmentation, we follow \citet{xie2019improving} to use simple random transformations so that different data patterns can be dug to avoid overfitting.
For model augmentation, we aim to obtain various models by modifying a single surrogate model and only focus on the back propagation process to ensure the model prediction keep the same.
Inspired by the SGM method~\cite{wu2020skip}, we introduce a set of learnable decay factors where one factor per layer to suppress the gradient from the residual branch.
Particularly,
we first optimize the decay factors according to the hypothesis that the ``linear nature'' of DNNs causes the adversarial vulnerability~\cite{goodfellow2015explaining,guo2020backpropagating},
and then randomly change them to generate diverse models.
After augmentation, we define attacking a combination of specific input and a modified model as a task,
and polish the perturbations iteratively on abundant tasks for better generalization.
We finally introduce meta-learning optimization to reduce the number of required tasks.
Unlike previous works~\cite{yuan2021meta,yin2021adv} performing meta learning on multiple surrogate models, we perform meta learning on augmented tasks.
To verify the effectiveness of our method,
we conduct extensive experiments on a widely used dataset, and achieve 12.85\% better attack performance compared with the state-of-the-art methods.
To further demonstrates the practical potentials of our method,
we evaluate it with baselines on the real-world online system, \emph{i.e.}, Google Cloud Vision API, where our method achieves 42\% attack success rate which is 5\% higher than the second-best one.

Concretely, we summarize our contributions as follows:
\begin{itemize}
\item We propose a novel transfer adversarial attack framework termed \emph{Learning to Learn Transferable Attack} (LLTA)
based on meta learning to generate generalized adversarial perturbations with data and model augmentation on a single surrogate model.
\item We propose an efficient strategy to enhance the transferability of adversarial perturbations by modifying model back propagation,
which can be extended as a manner of model augmentation.
\item Compared with the state-of-the-art methods, LLTA achieves a higher transfer attack success rate on ImageNet pre-trained models and real-world systems, demonstrating its effectiveness.
\end{itemize}

\section{Background and Related Work}
Given an input $\bm{x}$ with its label $y$ and a model $f(\cdot)$, the objective of adversarial attack can be formulated as:
\begin{align}
  \label{equ:optimization_objective}
  \max_{\bm{x}_{adv}}\ \mathcal{L}\big(f(\bm{x}_{adv}), y\big),
  \text{\ \ \ s.t.\ \ }
  \|\bm{x}_{adv} - \bm{x}\|_{p} \leq \epsilon,
\end{align}
where $\mathcal{L}(\cdot, \cdot)$ denotes the prediction loss (\emph{e.g.}, the cross-entropy loss),
and $\bm{x}_{adv}$ denotes the adversary example.
The $\ell_{p}$ norm of adversarial perturbation $(\bm{x}_{adv} - \bm{x})$ is constrained by the predefined value $\epsilon$,
and $\ell_{p}$ is set as $\ell_{\infty}$ in this paper following the most common setting~\cite{dong2018boosting}.

For transfer adversarial attack, with the full accessible surrogate model, white-box attack methods based on gradient are used.
The classic method, Fast Gradient Sign Method (FGSM)~\cite{goodfellow2015explaining}, crafts the adversarial example with a one-step update as:
\begin{align}
  \label{equ:fgsm}
  \bm{x}_{adv} = \bm{x} + \epsilon \cdot \text{sign}\Big(\nabla_{\bm{x}}\mathcal{L}\big(f(\bm{x}), y\big)\Big)
\end{align}
\citet{kurakin2016adversarial} extended FGSM to an iterative version (I-FGSM) with a small step size $\alpha$ as:
\begin{align}
  \label{equ:ifgsm}
  \bm{x}^{t+1}_{adv} = \Pi^{\bm{x}}_{\epsilon} \bigg(\bm{x}^{t}_{adv} + \alpha \cdot \text{sign} \Big(\nabla_{\bm{x}}\mathcal{L}\big(f(\bm{x}^{t}_{adv}), y\big)\Big)\bigg),
\end{align}
where $\Pi^{\bm{x}}_{\epsilon}(\cdot)$ ensures $\bm{x}_{adv}$ within the $\epsilon$-ball of $\bm{x}$.
I-FGSM improves the power of the white-box attack while impeding the transferability.
Based on it,
recent works aim to improve adversarial transferability, which can be divided into three categories: \emph{i.e.}, gradient optimization, input transformation, and model modification.

\textbf{Gradient Optimization:}
Some optimization techniques used to improve generalization have been introduced to improve adversarial transferability.
Momentum Iterative Method (MI)~\cite{dong2018boosting} integrates momentum~\cite{polyak1964some} into the gradient of each iteration, which allows previous gradients to guide the current updating direction to alleviate overfitting.
Instead of using the current gradient for momentum, \citet{2103.15571v1} tuned it with the gradient variance of the previous iteration.
\citet{lin2019nesterov} integrated Nesterov Accelerated Gradient~\cite{nesterov1983method} which is also a popular optimization method.

\textbf{Input Transformation:}
As an effective method for generalization, data augmentation has been utilized to make the perturbation underfit to specific data patterns.
Diverse Input Method (DI)~\cite{xie2019improving} applies different transformations (\emph{e.g.}, random resizing and padding) to the input image each iteration.
Translation Invariant Method (TI)~\cite{dong2019evading} aims to calculate gradients of a set of transformed images,
and applies a predefined kernel to the gradient according to translation invariance.
Scale Invariant Method (SI)~\cite{lin2019nesterov} optimizes perturbations over the scale copies of the input image.

\textbf{Model Modification:}
To avoid overfitting the specific surrogate model,
some works try to modify or enhance the surrogate model to a more general one.
Model ensemble~\cite{liu2017delving} is the most straightforward method via using multiple surrogate models,
and meta learning~\cite{yuan2021meta,yin2021adv} is utilized to further improve the performance.
However, multiple surrogate models are hard to obtain and extra computation is consumed.
More works tend to focus on only one surrogate model.
Ghost Networks~\cite{li_learning_2018} applies random dropout and skip-connection erasion to simulate plenty of models.
Intermediate Level Attack (ILA)~\cite{huang2019enhancing} crafts transferable adversarial examples by removing the later layers of the surrogate model
as low-level features extracted by DNNs may be shared.
Instead of altering the forward propagation process which influences predictions, some methods only modify the back propagation process.
\citet{wu2020skip} experimentally concluded the skip connection benefits adversarial transferability, and introduces a decay factor to reduce gradients from the residual modules.
Inspired by the linear nature hypothesis~\cite{goodfellow2015explaining},
Linear Backpropagation (LinBP)~\cite{guo2020backpropagating} skips some non-linear activation layers on back propagation.

\section{Methodology}

To improve the transferability of adversarial examples, we propose a novel framework termed \emph{Learning to Learn Transfer Attack} (LLTA),
 which generates generalized adversarial perturbations via meta learning with both data and model augmentation as depicted in Fig.\,\ref{fig:overview}.
By treating attacking a combination of a specific data and model as a task, we expect the crafted adversarial perturbations work on various tasks for transferability.
To this end,
we first propose a task generation module with both data and model augmentation as described in Sec.\,\ref{sec.task_generation},
and tasks are generated during optimization iteration.
Then we introduce the meta-learning strategy to make full use of generated tasks,
where tasks are divided into two sets and adversarial perturbations are crafted on one set and finetuned on another set (Sec.\,\ref{sec.meta-learning_optimization}).

\subsection{Task Generation}
\label{sec.task_generation}
The poor transferability is caused by the gap between the surrogate model and the victim model,
which can be further considered as the different data patterns extracted by two models and different behaviors of two models on forward and backward propagation.
To narrow the gap, we need to optimize the adversarial perturbation via both data and model augmentation in case of overfitting to the surrogate model.
Reviewing Eq.\,\ref{equ:ifgsm} in I-FGSM, the adversarial perturbations is updated mainly on the gradient, \emph{i.e.}, $\nabla_{\bm{x}^{t}_{adv}}\mathcal{L}\big(f(\bm{x}^{t}_{adv}), y\big)$.
We rewrite the gradient as $G(\bm{x}^{t}_{adv}, f)$ and define a task as attacking an augmented data and model.
Then, we can optimize the adversarial perturbation with tasks generated on the fly as:
\begin{align}
  \label{equ:task_ifgsm}
  & \bm{x}^{t+1}_{adv} = \Pi^{\bm{x}}_{\epsilon} \bigg(\bm{x}^{t}_{adv} + \alpha \cdot \text{sign}\big( \bm{g}^t \big) \bigg), \nonumber \\
  & \bm{g}^t = \Big(\frac{1}{n} \cdot \sum_{n} G\big(T_{d}(\bm{x}^{t}_{adv}), T_{m}(f)\big)\Big)\bigg),
\end{align}
where $T_{d}(\cdot)$ and $T_{m}(\cdot)$ denote random data transformation and model augmentation respectively.

\subsubsection{Data Augmentation}
Data augmentation is an effective method to enhance generalization and alleviate overfitting, which has been verified on various vision tasks.
As complex augmentation strategies are commonly non-differentiable which prevents us from optimizing perturbations by utilizing gradients,
we tend to seek simple and differentiable image transformations.
Following previous works~\cite{xie_mitigating_2018,xie2019improving}, we adopt the random resizing and padding operations.
Particularly, given an input image with the width and height of $W$, we first resize it to a random size $W' \in [W, W/0.875]$ following the common practice.
We then add random zeros pixels to the left and top of the resized image, and pad it on the right and bottom side for the size of $W/0.875$.
To align with the size of the initial image, we finally resize the padded image to the size of $W$.
Same as DI, we also transform the image with a probability of 0.5 to make sure the adversarial perturbation works for the original input.
Based on the above discussion, the data augmentation function can be represented as:
\begin{align}
& T_{d}(\bm{x}) = \nonumber
& \left\{
  \begin{aligned}
& \text{Resize}(\text{RandPad}(\text{RandResize}(\bm{x}))), & p \leq 0.5 \\
& \bm{x}, & p>0.5
  \end{aligned}
  \right.,
\end{align}
where $p \sim \mathcal{U}(0,1)$ controls to use transformed image or the original one.

\subsubsection{Model Augmentation}

Unlike model ensemble that needs multiple surrogate models,
we aim to modify only one surrogate model for model augmentation.
Moreover, to avoid perturbing the prediction of the pre-trained model, we focus on adjusting the back propagation process.
\citet{wu2020skip} has proposed the SGM method that introduces a decay factor in back propagation to reduce the gradient from the residual modules,
as the skip connections are widely used and the gradient from it is more helpful for transferability.
It inspires us to adopt different decay factors for model augmentation.

In SGM, only one single decay factor is used, which is chosen according to a validation set and is fixed during optimization for all input images.
The inflexible factor setting harms the efficiency of SGM, and hinders model augmentation.
As different layers play different roles and the decay factors should also vary for different inputs,
we introduce a set of changeable decay factors instead.
Denoting the set of decay factors as $\bm{\gamma}=[\gamma_{1}, \gamma_{2},...,\gamma_{L}]^{\mathrm{T}} \in [0,1]^{L}$ and the factor for $i$-th residual layer as $\gamma_{i}$,
we re-write the forward and back propagation of the residual layer as:
\begin{align}
  \label{equ:gradient}
  & \bm{z}_{i+1} = \bm{z}_{i} + \gamma_{i}\cdot f_{i+1}(\bm{z}_{i}) + C, \nonumber \\
  & G = \nabla_{\bm{x}}\mathcal{L} = \frac{\partial \mathcal{L}}{\partial \bm{z}_{L}} \prod^{L-1}_{i=0}(\gamma_{i+1} \frac{\partial f_{i+1}}{\partial \bm{z}_{i}}+1) \frac{\partial \bm{z}_{0}}{\partial \bm{x}},
\end{align}
where $\bm{z}_{i}$ denotes the input of $(i+1)$-th residual layer, $f_{i}$ denotes the $i$-th residual module,
and $C$ is a constant whose value is equal to $(1-\gamma_{i})\cdot f_{i+1}(\bm{z}_{i})$.
Such modification is easy to implement and brings nearly no extra computation cost.
Now, with the changeable decay factors, we can naturally augment the surrogate model by altering the value of factors.

Note that SGM achieves better results with a carefully picked decay factor,
which inspired us that well-tuned decay factors will bring further performance improvement.
To this end, we tend to sample the decay factors $\bm{\gamma}$ from an optimized distribution instead of a random one.
Here we optimized the $\bm{\gamma}$ to reduce the gradient w.r.t. inputs, \emph{i.e.}, $\| G \|_{2}$.
The design behind the objective is two-fold.
On the one hand, the $\bm{\gamma}$ should make the model towards locally linear for better transferability according to the ``linear nature'' hypothesis~\cite{goodfellow2015explaining,guo2020backpropagating}.
The linearity for a function $\ell(\bm{x})$ can be measured by the difference between it and its first-order Taylor expansion:
\begin{align}
  \label{equ:linear}
  & | \ell(\bm{x}+ \bm{\delta}) - \ell(\bm{x}) -  \bm{\delta}^{\mathrm{T}} \nabla_{\bm{x}} \ell(\bm{x})| \nonumber \\
  \leq
  & | \ell(\bm{x}+ \bm{\delta}) - \ell(\bm{x})| + | \bm{\delta}^{\mathrm{T}} \nabla_{\bm{x}} \ell(\bm{x})|.
\end{align}
Assuming the function is $K$-Lipschitz continuous, we can further bound the first term on the right side in Eq.\,\ref{equ:linear} as:
\begin{align}
  \label{equ:lipschitz}
  | \ell(\bm{x}+ \bm{\delta}) - \ell(\bm{x}) | \leq K |\bm{\delta}|.
\end{align}
Combining Eq.\,\ref{equ:linear} and Eq.\,\ref{equ:lipschitz}, we can conclude that reducing the gradient contributes linearity and transferability.

On the other hand, minimizing the gradient is also a classic method for robust model~\cite{jakubovitz_improving_2018} which hinders adversarial attack.
According to some regularization methods~\cite{wu2020adversarial,2010.04925v3},
optimizing in an adversarial situation brings a flatter loss landscape which actually introduces more generalization,
which also supports our design.

To optimize $\bm{\gamma}$,
we adopt the MLE-guided parameter search method (MGS)~\cite{welleck2020mle}.
MGS is a general optimization method regardless of whether the objective function is derivable or not,
and it introduces more uncertainty which suits the model augmentation compared with the simple gradient descent method,
In each iteration,
the MGS samples some update directions from a predefined distribution (\emph{e.g.}, a Gaussian distribution $\mathcal{N}(\bm{0}, \sigma^{2} \bm{\mathrm{I}})$),
mixes them using the corresponding improvement in the objective function as weight,
and finally update the parameter to be optimized.
This can be formulated as:
\begin{align}
    \label{equ:optim_update}
  & \bm{\Delta}_{\bm{\gamma}}^{n} \sim \mathcal{N}(\bm{0},\sigma^{2} \bm{\mathrm{I}}) \nonumber\\
  & p(\bm{\Delta}_{\bm{\gamma}}^{n} | \bm{\gamma}^{t}) = \frac{\exp \big(\|G(\bm{\gamma}^{t})\|_{2}-\|G(\bm{\gamma}^{t} + \bm{\Delta}_{\bm{\gamma}}^{n})\|_{2}\big)
      }{ \mathcal{N}(\bm{\Delta}_{\bm{\gamma}}^{n}; \bm{0},\sigma^{2} \bm{\mathrm{I}})}
      \nonumber \\
  & \bm{\gamma} = \bm{\gamma} + \sum \frac{p(\bm{\Delta}_{\bm{\gamma}}^{n} | \bm{\gamma}^{t})}{\sum p(\bm{\Delta}_{\bm{\gamma}}^{n} | \bm{\gamma}^{t})} \cdot \bm{\Delta}_{\bm{\gamma}}^{n}.
\end{align}
We then obtain an optimized $\bm{\gamma}^{\star}$ after several iterations.
Instead of performing MGS multiple times for multiple $\bm{\gamma}^{\star}$ leading high computational cost, we just optimize once, and then alter it randomly multiple times, \emph{i.e.}, add noises sampled from the uniform distribution on $[-0.5,0.5]^{L}$.
This achieves similar attack performance but saves a great deal of computation.
With multiple well-tuned decay factors, we finally complete model augmentation.

In summary, the model augmentation part can be stated as optimizing the decay factors $\bm{\gamma}^{\star}$ first, randomly changing them multiple times for multiple $\bm{\gamma}$,
and applying $\bm{\gamma}$ on the surrogate model separately.
As the $T_{m}(\cdot)$ is heavily about the $\bm{\gamma}$, we rewrite $G(\bm{x}, f)$ as $G(\bm{x}, \bm{\gamma})$.

\begin{algorithm}[!t]
\caption{Learning to Learn Transferable Attack (LLTA)}
\label{alg:LLTA}
\LinesNumbered
  \KwIn{
Source image $\bm{x}$, number of perturbation iterations $T$ and meta iteration $M$,
size of support set \#$\mathcal{S}$ and query set \#$\mathcal{Q}$.
}
\KwOut{Adversarial example $\bm{x}_{adv}$.}
    Initialize adversarial example $\bm{x}^{0}_{adv} = \bm{x}$. \\
    \For {$t=0$ \KwTo $T-1$} {
        \textbf{// Pre-process decay factors $\bm{\gamma}$} \\
        Initialize $\bm{\gamma}$ with $\bm{0.5}$. \\
        Optimize $\bm{\gamma}^{\star}$ by Eq.~\ref{equ:optim_update}. \\
        \textbf{// Create task sets} \\
        $\mathcal{S}=\left\{ \right\},\mathcal{Q}=\left\{ \right\}$ \\
        \For{$\mathcal{D} \in \{\mathcal{S}, \mathcal{Q}\}$} {
            \For{$i=0$ \KwTo \#$\mathcal{D}-1$}{
              $\mathcal{D} = \mathcal{D} \cup \big\{\big(T_{d}(\bm{x}_{adv}^{t}), \bm{\gamma}^{\star}+ \bm{\Delta}\big)\big\}$,
              $\bm{\Delta} \sim \mathcal{U}([-0.5,0.5]^{L})$. \\
            }
            }

        \textbf{// Meta-learning optimization} \\
        \For{$m=0$ \KwTo $M-1$}{
            Sample a mini-batch $\mathcal{S}_i$ from $\mathcal{S}$. \\
            \textbf{// Meta-Training} \\
            $\bm{g}_{spt} = \frac{1}{| \mathcal{S}_i |} \sum_{(\bm{x}_{s}, \bm{\gamma}_{s}) \in \mathcal{S}_{i}} G(\bm{x}_{s}, \bm{\gamma}_{s})$ \\
            $\bm{\delta'} = \epsilon \cdot \text{sign}(\bm{g}_{spt})$ \\
            \textbf{// Meta-Testing} \\
            $\bm{g}_{qry} = \frac{1}{| \mathcal{Q} |} \sum_{(\bm{x}_{q}, \bm{\gamma}_{q}) \in \mathcal{Q}} G(\bm{x}_{q} + \bm{\delta'}, \bm{\gamma}_{q})$ \\
        }
        $\bm{x}^{t+1}_{adv} = \Pi^{\bm{x}}_{\epsilon} \big(\bm{x}^{t}_{adv} + \alpha \cdot \text{sign} (\bm{\overline{g}}_{spt} + \bm{\overline{g}}_{qry})\big)$ \\
    }
    \Return $\bm{x}_{adv}^{T}$
\end{algorithm}

\subsection{Meta-Learning Optimization}
\label{sec.meta-learning_optimization}

Instead of simply generating the adversarial perturbation directly on these tasks,
we use the philosophy of meta-learning to optimize so that the perturbation can better fit various scenarios.
Unlike previous works~\cite{yuan2021meta,yin2021adv} which perform meta learning directly on multiple surrogate models,
we only have one surrogate model and perform meta learning on the tasks with augmentation.
Note that, the tasks we constructed are based on the diversity of transferable attacks,
so making the perturbation adapt these tasks better will not lead to overfitting.
The meta learning method is performed on each iteration of the perturbation update.
In each iteration,
with the tasks split into a support set and a query set,
we perform meta training (training on support set) and meta testing (finetuning on query set) multiple times,
and finally update the adversarial perturbations.

\subsubsection{Meta-Training}
With the tasks generation method described in Sec.\,\ref{sec.task_generation},
we can generate sufficient tasks and then split them into the support set $\mathcal{S}$ and the query set $\mathcal{Q}$.
In the meta-training stage, we use the support set $\mathcal{S}$ to craft a temporary adversarial perturbation.
In each meta learning iteration,
we sample a subset $\mathcal{S}_i \subset \mathcal{S}$
and follow Eq.\,\ref{equ:task_ifgsm} to calculate the average gradient w.r.t. input as:
\begin{align}
    \label{equ:spt_g}
    \bm{g}_{spt} = \frac{1}{| \mathcal{S}_i |} \sum_{(\bm{x}_{s}, \bm{\gamma}_{s}) \in \mathcal{S}_{i}} G(\bm{x}_{s}, \bm{\gamma}_{s}).
\end{align}
Then, similar as FGSM, we obtain the temporary perturbation with a single step update:
\begin{align}
    \label{equ:temp}
  \bm{\delta'} = \epsilon \cdot \text{sign}(\bm{g}_{spt}).
\end{align}

\subsubsection{Meta-Testing}
To improve the generalization of temporary perturbation, we finetune it on the query set $\mathcal{Q}$ so that it can adapt more tasks.
Particularly, we compute the query gradient $\bm{g}_{qry}$ here by adding the temporary perturbation:
\begin{align}
    \label{equ:qry_g}
    \bm{g}_{qry} = \frac{1}{| \mathcal{Q} |} \sum_{(\bm{x}_{q}, \bm{\gamma}_{q}) \in \mathcal{Q}} G(\bm{x}_{q} + \bm{\delta'}, \bm{\gamma}_{q}).
\end{align}

\subsubsection{Meta-Optimization}
With,
we finally update the actual adversarial perturbation with the gradient from both the support set and the query set for maximum utilization:
\begin{align}
    \label{equ:adv}
  \bm{x}^{t+1}_{adv} = \Pi^{\bm{x}}_{\epsilon} \big(\bm{x}^{t}_{adv} + \alpha \cdot \text{sign} (\bm{\overline{g}}_{spt} + \bm{\overline{g}}_{qry})\big),
\end{align}
where $\bm{\overline{g}}_{spt}$ and $\bm{\overline{g}}_{qry}$ denote the average gradient over meta learning iterations, respectively.

We refer to the method combining the task generation and meta learning optimization as Learning to Learn Transferable Attack (LLTA),
and summarize it in Alg.\,\ref{alg:LLTA}.

\section{Experiment}

\subsection{Experiment Setup}
\subsubsection{Datasets.}
Following most of the previous works, we report the results on the ImageNet-compatible dataset in the NIPS 2017 adversarial competition~\cite{kurakin2018adversarial},
which contain $1,000$ categories and one image per category.
We tune hyper-parameters on another $1,000$ images randomly chosen from ImageNet validation set~\cite{deng2009imagenet}.

\subsubsection{Surrogate and Victim Models.}
For surrogate models, following SGM~\cite{wu2020skip},
we choose eight different models from ResNet family~\cite{he2016deep} (Res-18/34/50/101/152) and DenseNet family~\cite{huang2017densely} (DenseNet-121/169/201).
We mainly report the results on the ResNet-50 and DenseNet-121 as they are more popular.

For victim models,
we consider two transferable attack scenarios, \emph{i.e.}, naturally trained models and robust models.
We pick popular models including above surrogate models along with VGG-16~\cite{simonyan2015very}, Inception-v3 (IncV3)~\cite{szegedy2016rethinking}, Inception-v4 (IncV4), and Inception-ResNet-v2 (IncResV2)~\cite{szegedy2017inception} for naturally trained models.
We will omit the white-box situation when the surrogate model and the victim model are the same.
And for robust models, following the common practice~\cite{dong2018boosting,xie2019improving,lin2019nesterov},
we choose four adversarially trained models~\cite{tramer2018ensemble}, \emph{i.e.},  IncV3$_{ens3}$ (ensemble of 3 IncV3 models), IncV3$_{ens4}$ (ensemble of 4 IncV3 models), IncResV2$_{ens3}$ and  IncV3$_{adv}$\footnote{\url{https://github.com/JHL-HUST/SI-NI-FGSM}}.

\subsubsection{Compared Baselines.}
To demonstrate the effectiveness of the our proposed LLTA,
we compare it with existing competitive baselines, \emph{i.e.} I-FGSM~\cite{kurakin2016adversarial}, DI~\cite{xie2019improving}, TI~\cite{dong2019evading}, MI~\cite{dong2018boosting}, SGM~\cite{wu2020skip} and LinBP~\cite{guo2020backpropagating}.
We report the transfer attack success rate as the evaluation metric, which measures whether the victim model keeps the top-1 prediction with the adversarial perturbation.

\subsubsection{Experiment Details.}
We follow the attack setting in most of previous works~\cite{kurakin2016adversarial, dong2018boosting, dong2019evading} with the maximum $\ell_{p}$-norm $\epsilon=16$, number of iteration $T=10$, and step size $\alpha=\epsilon/T=1.6$.
We set other parameters following the original setting in baselines.
LinBP applies the linearly backpropagate only on some specific layers,
and we follow their setting to choose layers for the ResNet family.
For DenseNet models which are not evaluated in their original paper, we test all layers and choose the layers for the best results on the validation set.
For our LLTA, we set the size of the support set as 20, the size of the query set as 10, and the number of meta iterations as 5, respectively.
For MGS used in LLTA, we set the number both of iteration and sampled update directions as 5.

\begin{table*}[t]
\centering
\resizebox{1\textwidth}{!}{
\begin{tabular}{l|l|c|c|c|c|c|c|c|c|c|c|c|c|c}
\hline
                                           & Attack        & VGG16         & RN18          & RN34          & RN50             & RN101         & RN152         & DN121            & DN169         & DN201         & IncV3         & IncV4         & IncResV2      & Avg.            \\
                                                                                                           \hline
                                           & I-FGSM        & {60.1}        & {63.9}        & {68.5}        & \textbf{ 100.0}*        & 79.6          & 69.2          & 68.5             & 63.8          & 61.7          & 30.7          & 25.5          & 19.5          & 55.55          \\
                                           & TI            & 59.7          & 66.7          & 70.4          & \textbf{ 100.0}* & 77.0          & 72.4          & 74.2             & 70.4          & 68.7          & 48.8          & 45.1          & 38.7          & 62.92          \\
                                           & DI            & 74.7          & 80.2          & 81.5          & \textbf{ 100.0}* & 86.7          & 81.8          & 81.2             & 80.0          & 77.1          & 49.6          & 44.7          & 36.3          & 70.35          \\
                                           & MI            & 80.0          & 84.1          & 87.0          & 99.9*            & 90.5          & 86.8          & 87.2             & 85.0          & 83.4          & 62.2          & 52.4          & 50.8          & 77.22          \\
                                           & LinBP         & 88.6          & 88.7          & 89.1          & \textbf{ 100.0}* & 93.6          & 89.1          & 90.1             & 87.7          & 85.5          & 55.6          & 56.5          & 48.5          & 79.36          \\
                                           & SGM           & 86.8          & 85.3          & 86.7          & \textbf{ 100.0}* & 92.8          & 88.2          & 86.1             & 85.2          & 83.4          & 55.0          & 54.1          & 45.4          & 77.18          \\
   \multirow{-7}{*}{\rotatebox{90}{RN50}}  & \textbf{LLTA} & \textbf{93.2} & \textbf{92.1} & \textbf{94.3} & 99.8*            & \textbf{95.5} & \textbf{93.5} & \textbf{92.9}    & \textbf{92.8} & \textbf{91.3} & \textbf{72.5} & \textbf{68.6} & \textbf{66.7} & \textbf{86.67} \\ \hline
                                           & I-FGSM        & {65.5}        & {66.5}        & {65.5}        & 73.2             & 63.9          & 56.9          & \textbf{ 100.0}* & 83.2          & 79.2          & 36.2          & 33.2          & 23.5          & 58.80          \\
                                           & TI            & 59.6          & 64.9          & 64.7          & 67.9             & 60.8          & 58.7          & \textbf{ 100.0}* & 79.2          & 78.1          & 52.2          & 45.8          & 40.7          & 61.15          \\
                                           & DI            & 74.8          & 75.8          & 78.6          & 81.0             & 73.0          & 72.2          & \textbf{ 100.0}* & 89.4          & 85.4          & 50.8          & 47.7          & 38.8          & 69.77          \\
                                           & MI            & 82.7          & 83.6          & 81.0          & 85.5             & 80.7          & 77.5          & \textbf{ 100.0}* & 91.9          & 90.4          & 64.1          & 59.3          & 53.6          & 77.30          \\
                                           & LinBP         & 91.5          & 87.7          & 86.8          & 90.7             & 83.6          & 81.5          & \textbf{ 100.0}* & 94.4          & 91.1          & 61.0          & 63.5          & 52.8          & 80.42          \\
                                           & SGM           & 84.5          & 84.5          & 85.9          & 88.3             & 83.5          & 81.9          & \textbf{ 100.0}* & 92.3          & 91.5          & 57.9          & 57.3          & 48.1          & 77.79          \\
   \multirow{-7}{*}{\rotatebox{90}{DN121}} & \textbf{LLTA} & \textbf{91.3} & \textbf{92.2} & \textbf{92.0} & \textbf{93.3}    & \textbf{90.5} & \textbf{89.7} & 99.9*            & \textbf{94.8} & \textbf{94.2} & \textbf{73.4} & \textbf{72.2} & \textbf{67.3} & \textbf{86.45} \\ \hline
\end{tabular}
}
\caption{The transfer attack success rate (\%) on naturally trained models. (RN: ResNet, DN: DenseNet, Avg.: average success rate on all black-box victim models, *: white-box attack.)}
\label{tab:natural}
\vspace{-1em}
\end{table*}

\subsection{Comparison of Transferability}
\subsubsection{Attacking Performance.}
We first evaluate the performance of our proposed LLTA and the baselines on naturally trained models.
We report the performance with ResNet-50 and DenseNet-121 as the surrogate model in Tab.\,\ref{tab:natural},
and report the average attack success rates on black-box victim models as well.
From Tab.\,\ref{tab:natural}, we can conclude that though with a little performance drop on the white-box scenario,
the adversarial perturbations generated by our method beat other methods on all black-box scenarios.
Averagely, our proposed LLTA achieves more than 86\% attack success rates for both ResNet-50 and DenseNet-121 models,
which are 6.67\% higher than LinBP and 9.07\% higher than SGM respectively.
We owe this to both data augmentation and model augmentation used in our method, which makes the adversarial perturbations more generalized.
We also find that the perturbations generated from ResNet-50 work better on ResNet models while worse on DenseNet models than perturbations from DenseNet-121.
It is consistent with the intuition that the transfer attack works better when the surrogate model is similar to the victim one.
Similarly, the performance of perturbations from both surrogate models degrades on Inception-based victim models as the models are more different.
However, for shallow models like VGG-16, both perturbations work well.

To further validate the performance of our method, we attack four adversarially trained models that are more robust,
and present the results in Tab.\,\ref{tab:robust}.
The average attack success rates degrade 30\%-40\% for all methods compared with naturally trained victim models.
Under this situation, our method achieves the best results with 46.18\% and 57.28\% success rates for two surrogate models,
which are more than 9.84\% higher than the second-best method MI.

\begin{figure}[!t]
    \centering
    \includegraphics[width=0.48\textwidth]{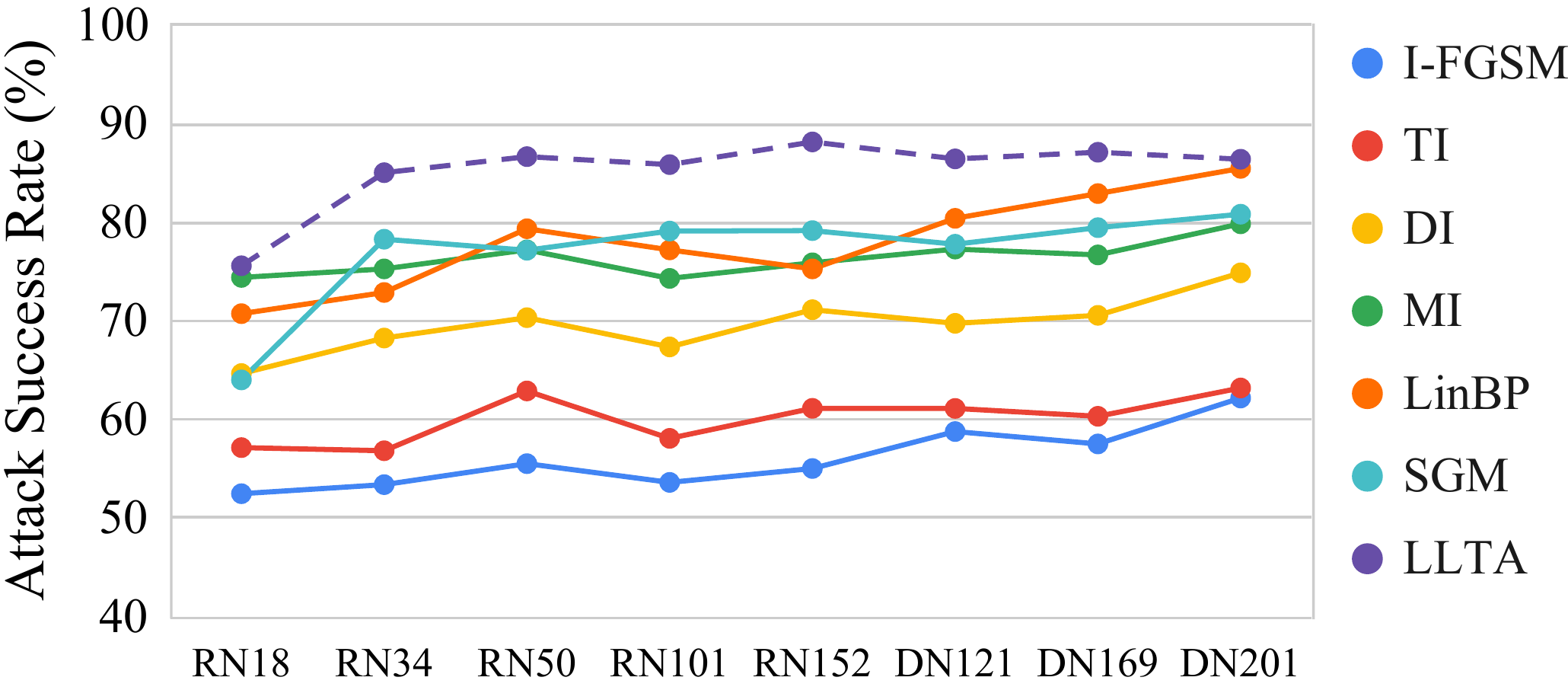}
    \caption{The average attack success rate on naturally trained models with various surrogate models.}
    \label{fig:surrogate_models}
\vspace{-1em}
\end{figure}

We report the average attack success on black-box scenarios for all eight surrogate models in Fig.\,\ref{fig:surrogate_models}.
Our method consistently outperforms baselines by a large margin for nearly all surrogate models.
Unlike methods like LinBP where the performance varies rapidly for different models,
our method keeps similar performance on different surrogate models except for ResNet-18,
which indicates that our method does not heavily depend on the surrogate model and the adversarial perturbations are generalized.
For ResNet-18 model,
as it is shallow and few decay factors are used (4 factors compared with 12 for ResNet-50),
the model augmentation is not fully utilized which leads to unsatisfactory performance.

\begin{table}[t]
\resizebox{0.48\textwidth}{!}{
\begin{tabular}{l|l|c|c|c|c|c}
\hline
                                         & Attack        & IncV3$_{ens3}$ & IncV3$_{ens4}$ & IncResV2$_{ens3}$ & IncV3$_{adv}$ & { Avg}            \\ \hline
                                         & I-FGSM        & 17.3           & 18.5           & 11.2              & 18.5          & { 16.38}          \\
                                         & TI            & 39.0           & 37.5           & 25.3              & 42.4          & { 36.05}          \\
                                         & DI            & 29.7           & 29.5           & 17.9              & 31.2          & { 27.08}          \\
                                         & MI            & 41.9           & 43.4           & 28.7              & 43.4          & { 39.35}          \\
                                         & LinBP         & 34.5           & 32.5           & 20.9              & 39.8          & { 31.93}          \\
                                         & SGM           & 30.4           & 28.4           & 18.6              & 36.9          & { 28.58}          \\
 \multirow{-7}{*}{\rotatebox{90}{RN50}}  & \textbf{LLTA} & \textbf{50.6}  & \textbf{47.3}  & \textbf{33.6}     & \textbf{53.2} & { \textbf{46.18}} \\ \hline
                                         & I-FGSM        & 21.8           & 21.5           & 13.1              & 23.5          & { 19.98}          \\
                                         & TI            & 40.4           & 41.2           & 28.2              & 47.6          & { 39.35}          \\
                                         & DI            & 30.6           & 32.0           & 20.5              & 35.2          & { 29.58}          \\
                                         & MI            & 46.4           & 48.6           & 33.8              & 48.9          & { 44.43}          \\
                                         & LinBP         & 39.3           & 38.3           & 22.6              & 44.5          & { 36.18}          \\
                                         & SGM           & 36.8           & 36.8           & 25.5              & 42.5          & {35.40}           \\
 \multirow{-7}{*}{\rotatebox{90}{DN121}} & \textbf{LLTA} & \textbf{59.1}  & \textbf{60.5}  & \textbf{46.8}     & \textbf{62.7} & { \textbf{57.28}} \\ \hline
\end{tabular}
}
\caption{Results on adversarially trained models.}
\label{tab:robust}
\vspace{-1em}
\end{table}

\subsubsection{Combining with Existing Methods.}
Considering that our method actually utilizes DI which may bring unfair comparison and
the performance of optimization methods like MI is competitive,
we finally report the average performance of LinBP, SGM, and our method combining with DI and MI on naturally trained models in Tab.\,\ref{tab:combination}.
We find that DI conflicts with LinBP that leads to large performance degradation, while promoting SGM with about 2\% performance improvement.
And the two methods are still not competitive with our methods with DI.
Combining with MI will bring improvement for all methods, and our method is still the best with at least a 5.32\% higher success rate than others.

\subsubsection{Various Epsilon $\epsilon$.}

\begin{figure}[!t]
    \centering
    \includegraphics[width=0.48\textwidth]{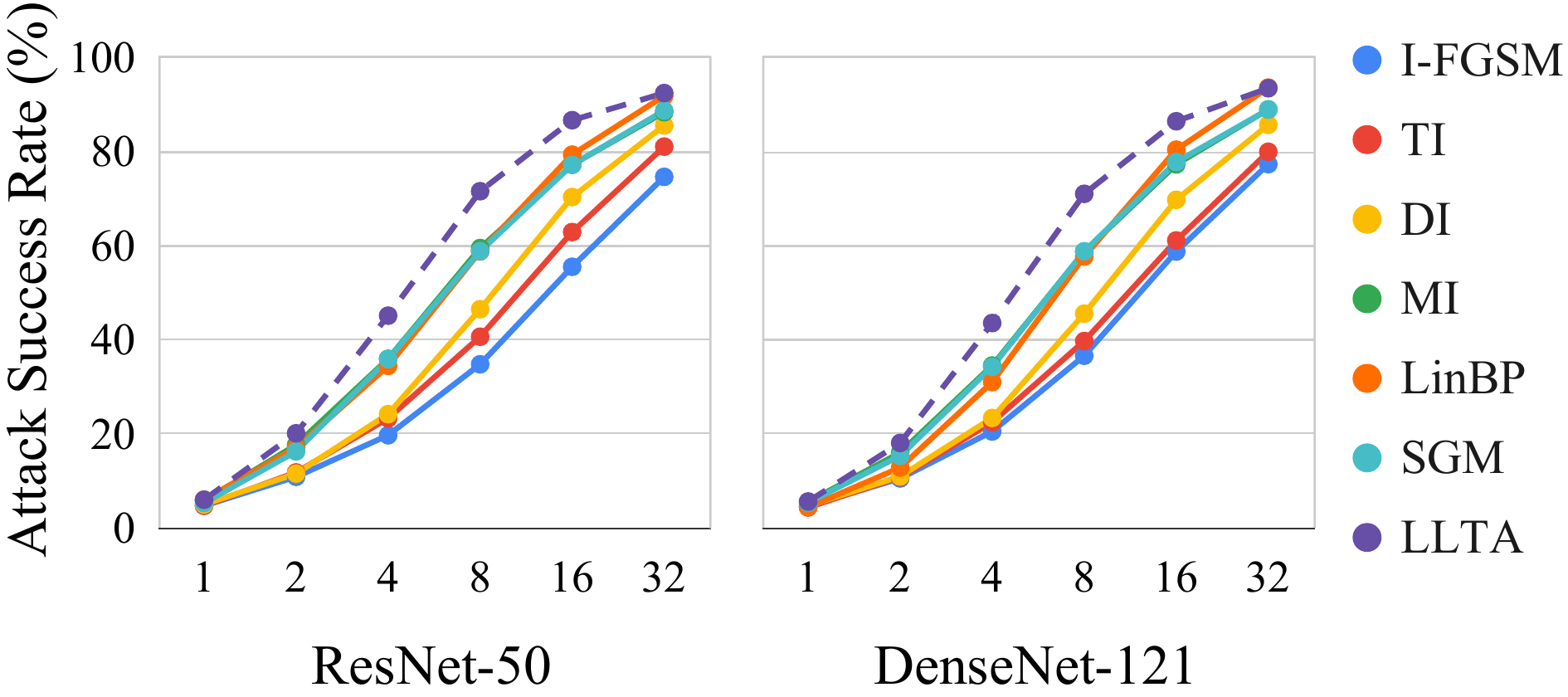}
    \caption{The average attack success rate on naturally trained models with various maximum perturbation $\epsilon$.}
    \label{fig:epsilon}
\vspace{-1em}
\end{figure}

We further evaluate the performance for attacking natural models under different epsilon $\epsilon$, \emph{i.e.}, $\epsilon=1,2,4,8,16,32$.
The results are depicted in Fig.\,\ref{fig:epsilon}.
The performance of all methods increases along with $\epsilon$ increases, and the curves are similar for ResNet-50 and DenseNet-121.
When the $\epsilon$ is small enough (\emph{e.g.}, $\epsilon=1$), all methods perform similarly and the success rates are all low.
Except this,
the curves of our method are above others with different values of $\epsilon$, which demonstrates the superiority of our method.

\begin{table}[t]
\centering
\begin{tabular}{m{2.1cm}|m{2.4cm}<{\centering}|m{2.4cm}<{\centering}}
\hline
   Attack       & ResNet-50      & DenseNet-121   \\ \hline
   LinBP        & 79.36          & 80.42          \\
   LinBP+DI     & 61.45          & 59.66          \\
   LinBP+MI     & 88.51          & 88.93          \\
   LinBP+DI+MI  & 81.12          & 79.05          \\
  \hline
      SGM       & 77.18          & 77.79          \\
      SGM+DI    & 79.46          & 79.57          \\
      SGM+MI    & 86.10          & 84.95          \\
      SGM+DI+MI & 88.31          & 87.76          \\
  \hline
      LLTA      & 86.67          & 86.45          \\
      LLTA+MI   & \textbf{92.24} & \textbf{91.76} \\
  \hline
\end{tabular}
\caption{The results of combining with DI and MI.}
\label{tab:combination}
\vspace{-1em}
\end{table}

\subsection{Ablation Study}
To verify the contribution of each part in LLTA, we conduct the ablation study by removing each part.
Our method relies on data augmentation and model augmentation along with the meta learning strategy.
Moreover, for model augmentation, we first train the decay factors for a better initialization, and then randomly alter them to obtain multiple models.
Based on the above discussion, we report the ablation results in Tab.\,\ref{tab:ablation}.
We first remove the data augmentation and model augmentation parts separately,
which brings 2.09\% and 23.60\% performance degradation, respectively.
It indicates that both data augmentation and model augmentation contribute the transferability, and model augmentation plays a more significant role.
It is easy to understand as data patterns are extracted from the model and model augmentation can bring diverse data patterns as well.
We further check each part in the model augmentation, \emph{i.e.}, optimizing decay factors and random alteration.
Without optimizing decay factors, the decay factors are all initialized as 0.5 and then randomly altered.
We observe a performance decline of more than 7.62\%.
It demonstrates well-tuned decay factors work better than random ones.
Without random alteration, it means only one augmented model with the optimized decay factors is used.
As it contributes to the model augmentation mainly, a poor attack result is obtained.
Finally, we evaluate the performance after removing the meta learning strategy and conclude that it is as significant as the random alteration.
In addition, we also verify whether using multiple decay factors instead of a single one as SGM is helpful.
We obtain an average success rate of 83.39\% with a single factor, and 84.83\% with 4 factors (one factor per block) respectively, compared with our method of 86.56\%.
Since both are lower than our method using one factor per residual layer and the result with 4 factors is higher than the one with 1 factor,
we conclude that using multiple decay factors does help for transferability.

\begin{table}[t]
\centering
\resizebox{0.48\textwidth}{!}{
\begin{tabular}{l|l|l|l|l}
\hline
                                           & Attack              & Natural            & Robust            & Total             \\ \hline
                                           & \textbf{LLTA}       & \textbf{86.67}     & \textbf{46.18}    & \textbf{75.87}    \\
                                           & -DataAugment        & 84.32  (\ \ -2.35) & 42.28 (\ \ -3.90) & 73.11 (\ \ -2.76) \\
                                           & -ModelAugment       & 57.95  (-28.72)    & 28.48 (-17.70)    & 50.09 (-25.78)    \\
                                           & \ \ -OptimizeFactor & 80.38  (\ \ -6.29) & 34.90 (-11.28)    & 68.25 (\ \ -7.62) \\
                                           & \ \ -RandomAlter    & 71.07  (-15.60)    & 38.35 (\ \ -7.83) & 62.35 (-13.53)    \\
   \multirow{-6}{*}{\rotatebox{90}{RN50}}  & -MetaLearning        & 74.43  (-12.24)    & 28.58 (-17.60)    & 62.20 (-13.67)    \\
  \hline
                                           & \textbf{LLTA}       & \textbf{86.45}     & \textbf{57.28}    & \textbf{79.11}    \\
                                           & -DataAugment        & 86.05  (\ \ -0.40) & 53.10 (\ \ -4.17) & 77.70 (\ \ -1.41) \\
                                           & -ModelAugment       & 63.77 (-22.68)     & 34.55 (-23.73)    & 57.69 (-21.41)    \\
                                           & \ \ -OptimizeFactor & 81.01  (\ \ -5.44) & 42.38 (-14.90)    & 71.37 (\ \ -7.74) \\
                                           & \ \ -RandomAlter    & 72.66  (-13.79)    & 44.55 (-12.73)    & 66.38 (-12.73)    \\
   \multirow{-6}{*}{\rotatebox{90}{DN121}} & -MetaLearning        & 75.96  (-10.49)    & 37.88 (-19.40)    & 66.73 (-12.38)    \\
  \hline
\end{tabular}
}

\caption{The result of albation experiments: w/ complete LLTA (line $1$); w/o data augmentation (line 2); w/o model augmentation (line 3); w/ decay factors initialized as 0.5 and not optimized (line 4); w/ decay factors optimized w/o further random alteration (line 5); w/o meta learning (line 6).}
\label{tab:ablation}
\end{table}

\subsection{Results on Real-world System}

\begin{figure}[!t]
    \centering
    \includegraphics[width=0.49\textwidth]{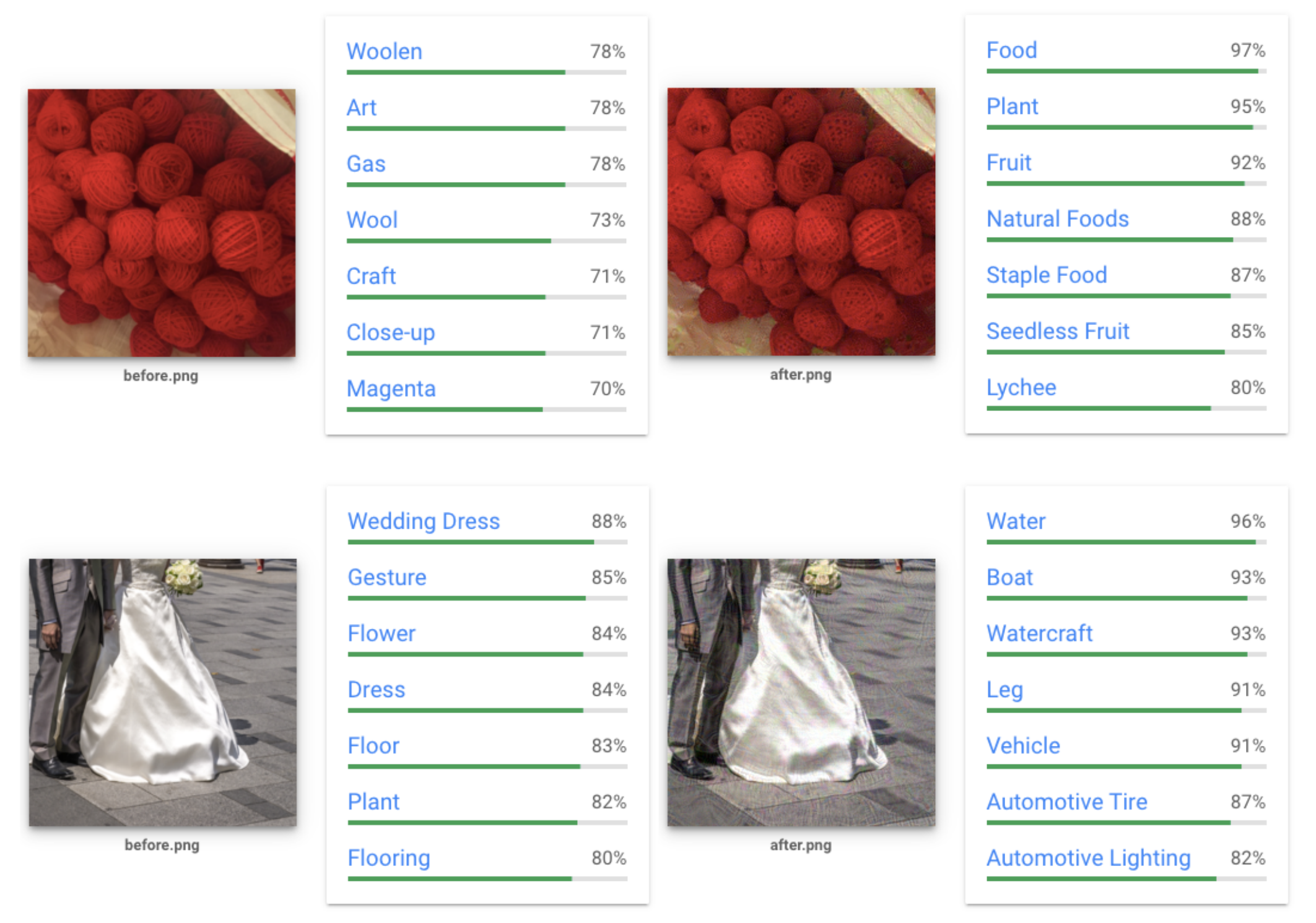}
    \caption{Examples for attacking Google Cloud Vision API. Images on the left are the original ones while images on the right are perturbed by LLTA on ResNet-50.}
    \label{fig:google}
\vspace{-1em}
\end{figure}

To explore the practical potential of our method, we finally test the adversarial examples generated by LLTA along with SGM and LinBP on a real-world image recognition system, \emph{i.e.}, Google Cloud Vision API\footnote{\url{https://cloud.google.com/vision/docs/drag-and-drop}}.
Given an image, this API will return a list of probable labels.
We randomly select 100 images from the dataset and generate adversarial examples with ResNet-50 as the surrogate model for evaluation.
We define a successful attack as the original top-1 prediction not appearing in the return list of the adversarial image,
and get 22\%, 37\%, and 42\% success rates for SGM, LinBP, and our method, respectively.
We also give two visualization examples in Fig.\,\ref{fig:google}, where woolen is recognized as fruit and wedding dress is recognized as boat above water.

\section{Conclusion}

In this paper, we track the problem of adversarial perturbations overfitting to the surrogate model in the transfer adversarial attack.
we propose a novel framework termed Learning to Learn Transferable Attack (LLTA),
which generates generalized adversarial perturbations by meta learning on tasks including both data and model augmentation.
In particular, we first define a task as attacking a specific data and model and generate tasks via data and model augmentation.
For data augmentation, simple random resizing and padding are utilized.
For model augmentation, a set of changeable decay factors is introduced.
Finally, we use the meta learning method to further optimize the adversarial perturbations over tasks.
We evaluate our proposed method on the widely-used ImageNet-compatible benchmark with popular pre-trained neural networks,
and show a higher transfer attack performance of LLTA compared with the state-of-the-art methods.
We also check the practicality of our proposed method against the real-world image recognition service, \emph{i.e.}, Google Cloud Vision API.
We hope this work can serve as an inspiration for more generalized methods and robust models.
We also believe that the decay factors learned by our method can help interpretability for deep models.
Besides, LLTA consumes more time due to extra iterations for optimal $\bm{\gamma}$ and meta learning, which is unsuitable for the online scenario.
We leave these for future work.

{\bibliography{aaai22.bib}}

\end{document}